# Differentiable Disentanglement Filter:
# an Application Agnostic Core Concept Discovery Probe

**Guntis Barzdins**
University of Latvia, IMCS
Rainis Blvd 29
Riga, Latvia
guntis.barzdins@lumii.lv

**Eduards Sidorovics**
University of Latvia
Rainis Blvd 19
Riga, Latvia
e.sidorovics@gmail.com

## Abstract

It has long been speculated that deep neural networks function by discovering a hierarchical set of domain-specific core concepts or patterns, which are further combined to recognize even more elaborate concepts for the classification or other machine learning tasks. Meanwhile disentangling the actual core concepts engrained in the word embeddings (like word2vec or BERT) or deep convolutional image recognition neural networks (like PG-GAN) is difficult and some success there has been achieved only recently. In this paper we propose a novel neural network nonlinearity named Differentiable Disentanglement Filter (DDF) which can be transparently inserted into any existing neural network layer to automatically disentangle the core concepts used by that layer. The DDF probe is inspired by the obscure properties of the hyper-dimensional computing theory. The DDF proof-of-concept implementation is shown to disentangle concepts within the neural 3D scene representation – a task vital for visual grounding of natural language narratives.

## 1 Introduction

The recent success with disentangling the semantically meaningful core concept dimensions within the representations learned by the popular deep neural networks (Dupont, 2018; Subramanian et al., 2018; Shabo 2018; Hewitt et al. 2019; Locatello et al., 2019) reveals that the "black box" un-interpretable nature of the neural networks is not their inherent property, but rather a byproduct of the too relaxed constraints during their training.

In this paper we introduce a Differentiable Disentanglement Filter (DDF) nonlinearity which can be transparently inserted into any existing deep neural network layer to disentangle the actual core concepts used by that layer. The approach is inspired by the obscure properties of the hyper-dimensional computing theory (Kanerva, 2009) developed before the current deep neural network revolution. This paper is the first proof-of-concept implementation on the DDF idea (Barzdins, 2018) and more rigorous testing in other domains including NLP is still ongoing.

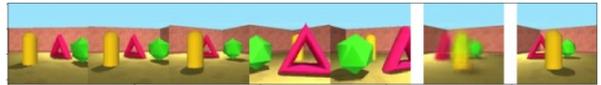

Figure 1: Given five views of the same 3D scene (left) the GQN reconstructs a view from any other viewpoint (center). Ground truth (right).

The DDF implementation in this paper is tested on disentangling only one, albeit a rather famous GQN deep neural network by Eslami et al. (2018) of high importance to both computer vision and NLP communities (Hermann et al., 2017; Barzdins et al., 2017). The generative GQN is chosen also for the clear visualization of the DDF results (Figure 1). The currently reported results provide only a low-level disentanglement – a stepping-stone towards a complete understanding of the representations learned by the GQN or other deep neural networks.

The paper is organized as follows: Section 2 discusses the related work on the semantic disentanglement of the deep neural network dimensions with emphasis on NLP, Section 3 introduces hyper-dimensional computing and DDF, Section 3 describes the GQN and DDF implementation used and Section 4 concludes with the DDF results on the GQN reference implementation.

## 2 Related Work on Disentanglement

Since the early days of connectionism (Hinton et al., 1986) it has been speculated that the concepts we use in the natural language are not entirely independent from each other, but rather are points distributed in the multidimensional space where some concepts are closer to each other while others are further apart. Widespread success of neural word embeddings (Mikolov et al., 2013) further cemented this distributional view about the nature of the concepts while also implicitly revealing the compositional structure of the embedding space with word analogies like *King-Man+Woman≈Queen.* Sparse non-negative embeddings (Subramanian et al., 2018) allow disentangling the semantic dimensions (core concepts) of the word embedding space at the cost of slightly increasing the number of the embedding dimensions. Unlike indirect compositionality evidence from the word analogies, the sparse non-negative embedding disentanglement provides an explicit word meaning decomposition into the core concepts represented by the semantically clear embedding dimensions. This disentangling success was a direct stimulus for this paper; we also observed that the semantic clarity of the dimensions greatly improves after the normalization of the embedding vectors. The non-negative aspect of this embedding disentanglement is reused also in our DDF approach.

Moving beyond words towards whole sentence embeddings has been more challenging for disentanglement (Conneau et al., 2018). Despite the early success of seq2seq embedding models for neural machine translation (Bahdanau et al., 2014), only the recent BERT embeddings (Devlin et al., 2019) have enabled partial sentence level embedding disentanglement (Hewitt et al., 2019). Disentanglement of the BERT sentence level embeddings with the linear transformation (learned in the supervised manner) revealed the full dependency parsing tree encoded by part of the dimensions. This result shows that the disentangled dimensions might have vastly different functions (what we observe also in our results in Section 4) with only a small fraction of the dimensions involved with the encoding of the parsing tree graph. However, contrary to the post-processing and supervised approach to disentanglement used in (Hewitt et al., 2019), our DDF approach performs disentanglement in completely unsupervised manner during the training of the host deep neural network.

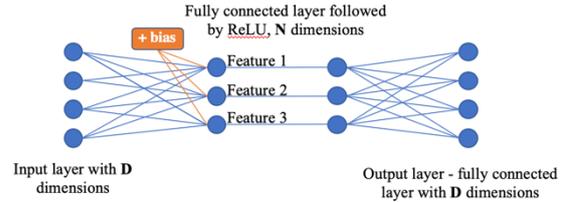

Figure 2: Differentiable Disentanglement Filter (DDF)**.**

Other examples of successful semantic dimension disentanglement in the visual domain are Dupont, (2018), Shabo (2018) and Locatello et al., 2019.

## 3 Differentiable Disentanglement Filter

We propose a novel semi-transparent and differentiable neural network nonlinearity called Differentiable Disentanglement Filter (DDF). Due to its multiple inputs and outputs the DDF externally resembles the popular Softmax nonlinearity, but internally DFF is organized like a small neural network (Fig. 2). The DDF consists of two fully connected linear neuron layers and a ReLU nonlinearity layer between them. What makes DDF to act as a regular nonlinearity function is that it is differentiable during backpropagation just like any regular neural network, but the weights within the fully connected layers of DDF are never updated during the training; thus the DDF as a function is defined by the initial random initialization of its weights.

The reason why such DDF nonlinearity is able to disentangle the semantic dimensions is explained by the hyper-dimensional computing theory (Kanerva, 2009). This theory builds upon the fact that high-dimensional vectors randomly initialized with uniform bipolar weights are mutually nearly orthogonal in the sense that their dot product is nearly 0. The dot product can deviate from 0 only if the two involved vectors correlate. If one of the involved random vectors A is unknown, it is impossible to build another vector B which would correlate with A. Meanwhile the sum of random vectors A+B correlates with both vectors A and B allowing to encode their compositionality within a single vector.

This hyper-dimensional computing theory suggests two easy to implement constraints on the weight initialization within the DDF to make it an orthogonal vector disentangling bottleneck: the random initialization of the weights must be bipolar (roughly the same number of weights must be positive and negative) and the bias weight must be negative. The first constraint is fulfilled by the popular normalized Gaussian, Xavier or Kaiming (Glorot, and Bengio, 2010; He et al., 2015) random initializations. Meanwhile the negative bias weights effectively regulate the ReLU nonlinearity cut-off point and should be set above the noise level to recognize only the true correlation between the DDF input activations and input weights. In practice we set the bias weight to small negative random values.

The randomly initialized output layer of the DDF again linearly entangles the orthogonal features detected by the DDF hidden layer and changes the output dimensionality to match the input dimensionality.

The DDF nonlinearity is fully differentiable and semi-transparent (it has equal number D of input and output neurons) and thus can be inserted between any layers within the existing DNN as a probe. DDF nonlinearity has one hyperparameter N – the number of hidden ReLU neurons. In our experiments we set N = D = 256 (the size of the representation layer within the host GQN network). To avoid disentangled feature duplication, ideally N would need to be adjusted to the lowest possible value with which the host network is still retrainable to its original quality.

Kanerva (2009) in his seminal paper also proposed a way to encode an arbitrary graph inside a high-dimensional vector. The scalar multiplication C=A*B of random vectors A and B is orthogonal to both A and B thus allowing to conceal the "entanglement" of vectors A and B inside C. Scalar multiplication of C with any of its components untangles the other component allowing to encode a key-value pair in a high-dimensional vector. By combining sum and scalar multiplication operations a set of graph vertex pairs (graph) can be encoded in a single high-dimensional vector. We did not include this graph-encoding option in the current incarnation of the DDF, but mention it as an interesting exploration path in the future.

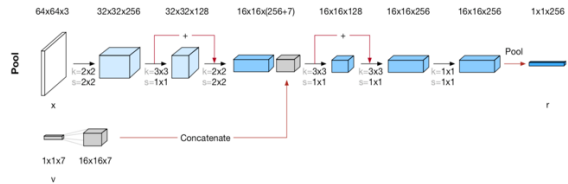

Figure 3: Representation network (Pool architecture).

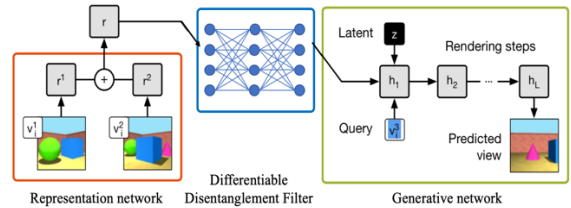

Figure 4: GQN-DDF architecture.

## 4 GQN and DDF Implementation

For the proof-of-concept test of DDF we rely on the Generative Query Network (GQN) introduced by Eslami et al. (2018). This model learns 3D scene representation from 2D input views (images and information about the viewpoint angle and location) and predicts views from previously unobserved viewpoints (Fig. 1).

In their seminal paper Eslami et al. (2018) experimented with several architectures for representation part of the model. In this paper we use Pool architecture – convolutional network with Global Average Pooling layer at the end (Fig. 3). Even though Eslami et al. noted that this architecture was not the fastest to learn across datasets, it was more likely to exhibit view-invariant. Moreover, Pool architecture provides a more convenient shape (1x1x256) for the initial DDF implementation testing.

To get the final representation tensor, the observed view goes through the chosen representation network and is concatenated with a reshaped vector of viewpoint parameters (Fig. 4). If multiple views are given, the final representation tensor is a sum of all single representation tensors.

For the generative part of the GQN Eslami et al. (2018) suggest employing a state-of-the-art deep, iterative, latent variable density model – Convolutional DRAW (Gregor et al., 2016). With this model, it is possible to generate views accounting for objects, background, lighting, shadows etc. Since Convolution DRAW model is probabilistic, GQN can handle uncertainty and produce images even in case of occlusion or partial observability. Our implementation of GQN model (Eslami 2018 provided only the training data, not the actual code) is written within the PyTorch framework and is available at https://github.com/esidorovics/gqn-pytorch.git.

Eslami et al. noted that "The values of all hyper-parameters were selected by performing informal search. We did not perform a systematic grid search owing to the high computational cost". Hence, we deviated from the original hyper-parameters and used ones recommended in the author's later publications on GQN (Kumar et al., 2018).

One of the key hyper-parameters for the Convolution DRAW is the number of generative steps. Gregor et al. (2016) showed that more generative steps produce sharper images and Eslami et al. (2018) suggest using 12 generative steps to train GQN model. However in our experiments we have used 8 steps because: (1) training takes less time, (2) produced results are still of reasonable quality. 8 generative steps is a good balance between the training speed and quality.

In our experiments we are interested in disentangling representation of the 3D space generated by the GQN model, hence DDF layer was integrated between the representation and generative parts of the model. General architecture can be observed in Figure 4. DDF layers were initialized by Kaiming (Kaiming et al., 2015) random initialization. Due to high computational costs we have experimented only with N=256 (size of the hidden bottleneck layer of the DDF) which is equal to the GQN representation layer size 1x1x256 as shown in Figure 3.

## 5 Experimental Results

### 5.1 Training process

We have trained GQN-original and GQN-DDF networks for 400,000 iterations with the same hyper-parameters (the only difference being DDF inserted after the representation calculation into one of them). Such training takes about one week on the Intel i7-5820K 6-core CPU workstation with 64GB DRAM and Nvidia GeForce GTX TITAN X, 12GB GPU. GQN end-to-end training loss consists of two terms: reconstruction loss and KL divergence. Reconstruction loss behaved identically, while KL divergence initially was slower to pick up (Figure 5) for GQN-DDF. Nevertheless, DDF effect on the training process was negligible.

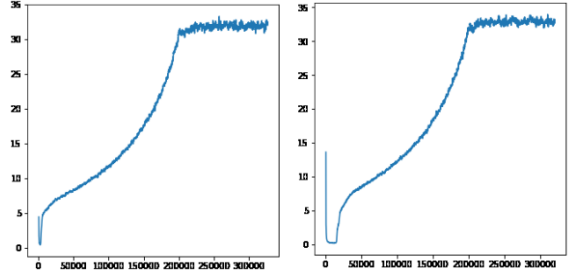

Figure 5: KL Divergence at training GQN-original (left) and GQN-DDF (right). DDF layer presence has negligible effect on the training speed and quality.

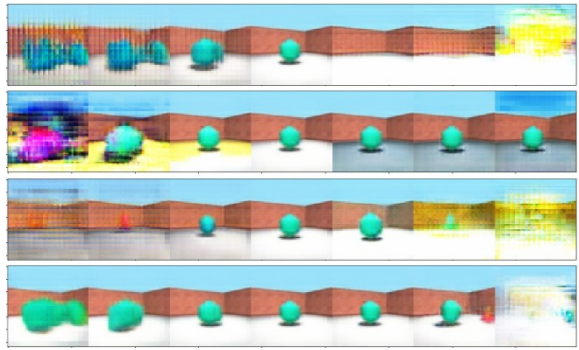

Figure 6: The 3D scene in the GQN-original is represented by the non-disentangled 256 neuron vector (middle image). Increasing (right) or decreasing (left) the value of individual neurons distorts all features of the image.

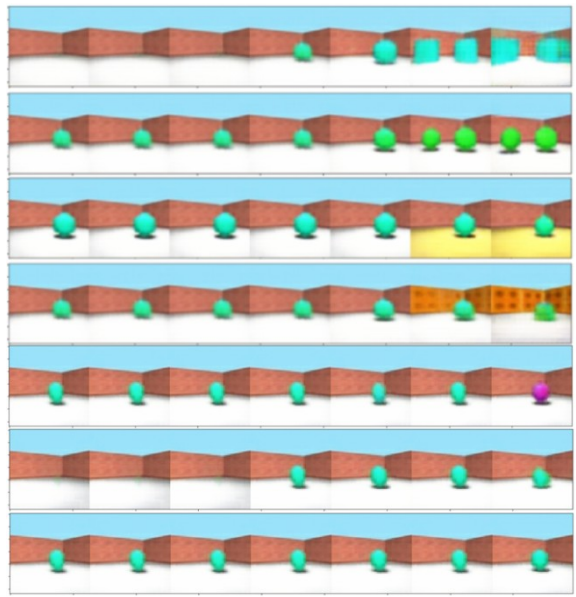

Figure 7: The 3D scene in the GQN-DDF is represented by the disentangled 256 neuron vector (middle image). Increasing (right) or decreasing (left) the value of individual neurons changes individual features (floor, wall, shaped object, two objects).

## 5.2 GQN-DDF and GQN-original comparison

Results of image reconstructions from both GQN-original and GQN-DDF were of similar quality. Differences in the produced details can be attributed to the probabilistic nature of the generative model. Hence the DDF insertion did not affect the eventual GQN results.

The main purpose of this paper is to test if the DDF layer can disentangle the concepts from the complex 3D representation vector and if the DDF disentanglement is superior to the original GQN representation. To test these hypotheses, we went through all 256 neurons of the hidden layer in DDF and manually changed individual neuron value by [-0.3, -0.2, -0.1, 0, +0.1, +0.2, +0.3] and observed how the generated image differed from the image without any alterations (Fig. 7). To test disentanglement level of GQN-original and to compare it with GQN-DDF we performed the same manipulations with the GQN-original representation layer neurons as well (Fig. 6)

Figures 6 and 7 summarize the results of such manipulations with the representation layers of GQN-original and GQN-DDF respectively. We will discuss differences and similarities from the two viewpoints – the image quality and the level of disentanglement.

**Image quality:** middle images in Figures 6 and 7 are quite similar; difference in quality can be attributed to the probabilistic nature of the model. However, the farther we go away from the "true" value of the neuron the higher the distortions are in the GQN-original model. In the GQN-DDF model quality also decreases, but to a much lesser extent. One of the obvious reasons is ReLU presence in the GQN-DDF model; however it explains only the lack of distortion on the left side suggesting the positive impact of the DDF layer.

**Disentanglement**: In Figure 6 we can observe that change in the single neuron of GQN-original representation triggers multiple feature changes in the generated image - size, color, shadows, wall texture etc. This identifies correlation between neurons, hence no disentanglement. In case of GQN-DDF, changes in some neurons also trigger multiple feature changes, however to much lesser extent (Figure 7). We have observed that the number of single feature neurons (change of single neuron value affects only one feature in the generated image) have been increasing over training, which makes us conclude that DDF will continue learning to improve disentanglement of the complex 3D representations.

## 5.3 Disentanglement results

Observation of produced DDF disentanglement results (Figure 7) makes us distinguish 3 types of disentanglement: (1) continuous, (2) discrete and (3) redundant.

1. Continuous: In the first two rows the value of the neurons affects the output. In the first row, ball is absent if neuron value is decreased. ReLU blocks us from observing what the negative value would have done with the output image. But increasing the value changes color, adds an additional object and modifies the shape of the object. Similar behavior is observed in the second row – increase in value adds a new object and changes the color. Even though it cannot be called a clear disentanglement because multiple features are affected by the single neuron, it seems to be limited to just a few features and to decrease with more training.
2. Discrete: Next four layers represent clear disentanglement. A certain neuron from a specific threshold value represents a specific feature: yellow floor, orange walls, purple color of the object. Also, interesting to note that the neuron in the 6$^{th}$ row represents the object itself, since decreasing value will make the object disappear. If we increase the value – the object stays the same. Disentanglement can be called discrete as a certain value serves as a "switch" for a specific feature, but the value of the neuron (after the specific threshold value) has no effect whatsoever.
3. Redundant: Last row in Figure 7 shows that changing a specific neuron has no effect in the generated image. This result indicates that the current neuron represents no feature and is uncorrelated with other neurons, hence the number of dimensions in the hidden layer of DDF could be reduced or with more training some feature might disentangle and "move" to this neuron.

## 6 Conclusions

The DDF layer design described in this paper might seem counter-intuitive and therefore more rigorous testing of the DDF properties has to continue, especially in other domains like NLP. But the ReLU nonlinearity or dropout regularization were also counter-intuitive before they were proved to actually work well and become the staple of the modern deep neural networks.

Successful disentanglement of the core concepts in each layer is only the first step towards understanding the internal logic of the deep neural network. Nevertheless, it is a vital first step towards such deciphering. It shall also be noted that the disentangled semantic dimensions appear to be somewhat arbitrary as observed already in Subramanian et al. (2018) – although we typically see individual objects, their color, size and shape getting disentangled, there remain other dimensions with less clear semantics which are either affected by the specifics of the domain (training data set) or the representation techniques beyond our current understanding.